\documentclass[a4paper,fleqn]{cas-dc}

\usepackage[numbers]{natbib}
\usepackage{microtype}
\usepackage{stfloats}
  
\def\tsc#1{\csdef{#1}{\textsc{\lowercase{#1}}\xspace}}
\tsc{WGM}
\tsc{QE}

\begin{document}
\let\WriteBookmarks\relax
\def\floatpagepagefraction{1}
\def\textpagefraction{.001}

\shorttitle{}    

\shortauthors{}  

\title [mode = title]{MRI-to-CT Translation of Spinal Images via Anatomy-Guided Pseudo-Pairing and Adversarial Diffusion}  



%

\author[1]{Eduardo Nieto}






\affiliation[1]{organization={Fundación Vicomtech, Basque Research and Technology Alliance (BRTA)},
            city={Donostia - San Sebastián},
            country={Spain}}

\author[2]{Ibai Genua}





\affiliation[2]{organization={R\&D Department, Cyber Surgery},
            city={Donostia - San Sebastián},
            country={Spain}}

\author[2]{Davide Scorza}





\author[1,3]{Karen López-Linares}
\cormark[1]

\ead{klopez@vicomtech.org}



\affiliation[3]{organization={eHealth Group, Bioengineering Area, Biogipuzkoa Health Research Institute},
            city={Donostia - San Sebastián}, 
            country={Spain}}

\cortext[1]{Corresponding author}



\begin{abstract}
\mbox{}\textit{Background:} Magnetic Resonance Imaging (MRI) is routinely used for diagnosing spinal disorders, but surgical planning often still requires Computed Tomography (CT) because of its superior visualization of vertebral bone anatomy. Avoiding an additional CT acquisition would reduce ionizing radiation exposure and simplify the preoperative workflow. MRI-to-CT synthesis therefore offers a promising route to provide CT-like anatomical information directly from MRI. \\

\noindent\textit{Methods:} We propose a three-stage framework for MRI-to-CT translation in spine imaging. First, a vertebral matching system uses spine segmentations to identify anatomically similar MRI and CT sagittal slices and align them by rigid registration, generating pseudo-paired training data from unpaired datasets. Second, the pseudo-paired slices are used to train an adversarial diffusion model for MR-to-CT synthesis. Third, the translated sagittal slices are reassembled into 3D volumes and refined with a 3D post-processing network designed to improve inter-slice consistency and volumetric structural coherence. \\

\noindent\textit{Results:} The approach was trained on more than 10,000 unpaired sagittal MR and CT slices and evaluated on 1,911 sagittal slices from 83 MR volumes covering all spinal regions. Evaluation included distribution-based metrics and task-specific vertebral segmentation metrics. Compared with the diffusion-only baseline, the proposed strategy improved vertebral reconstruction, increasing the binary 3D Dice score from 0.666 to 0.739. The 3D post-processing network reduced visible inter-slice discontinuities after sagittal slice stacking and produced more coherent vertebral structures across axial and coronal views, supporting improved volumetric consistency of the synthetic CTs. \\

\noindent\textit{Conclusions:} The proposed framework improves spine MRI-to-CT synthesis by combining anatomy-guided pseudo-pairing with adversarial diffusion-based translation and volumetric refinement. Vertebral matching was the main driver of improved anatomical alignment and vertebral fidelity, while post-processing enhanced 3D coherence. These results support the potential of the method as a step toward CT-like spine visualization from MRI for preoperative planning, with future work focused on preserving fine intensity detail while further strengthening volumetric consistency. 
\end{abstract}





\maketitle

\section{Introduction}
\label{sec:introduction}
Medical imaging modalities such as Magnetic Resonance (MR) and Computed Tomography (CT) provide complementary anatomical information of the spine region. MR offers superior soft-tissue contrast, enabling detailed visualization of intervertebral discs, the spinal cord, nerves, and surrounding soft tissues, whereas CT is particularly effective for depicting osseous structures such as vertebrae \citep{dhawan2011medical}. In spine surgery and preoperative planning, the combined use of both modalities enables accurate vertebral delineation, precise localization of critical neural structures, and detailed characterization of vertebral morphology, all of which are essential for decision-making \citep{harada2020imaging, oulbacha2020mri}.

In clinical practice, MR is generally the preferred modality for assessing most spinal pathologies, including disc herniation, spinal cord lesions, and tumors, because it provides rich soft-tissue information without ionizing radiation. CT is typically reserved for applications requiring detailed evaluation of bone anatomy, such as fracture assessment, implant planning, and surgical navigation, despite the associated radiation exposure. This complementary but selective use of MR and CT motivates the development of MR-to-CT translation methods, which aim to provide CT-like bone information directly from MR acquisitions.

Recent advances in deep learning have made this objective increasingly feasible by learning cross-modal mappings between MR and CT images \citep{dayarathna2024deep}. However, most existing MR-to-CT translation methods operate exclusively on 2D slices due to computational constraints, which limits their ability to enforce inter-slice consistency and full 3D anatomical coherence. Furthermore, a significant challenge in training these models is the scarcity of large, representative paired datasets \citep{mcnaughton2023machine}. Even when paired data are available, alignment between both imaging studies can be compromised by respiratory motion, tissue deformation or differences in patient positioning during acquisition \citep{bajger2021lumbar, vallejo2024cyclegan}. Given the higher availability of unpaired datasets, unsupervised methods have emerged as a viable alternative, showing promising results in medical image translation tasks \citep{wolterink2017deep, paavilainen2021bridging}. 

Thus, in this work, we propose a three-stage framework for MR-to-CT translation of spine images. First, a 2D vertebral matching system is implemented to generate pseudo-paired MR-CT data from different patients. These pseudo-pairs are then used to train a 2D SynDiff architecture \citep{ozbey2023unsupervised} across all spinal regions, enabling robust cross-modal synthesis despite the lack of true paired acquisitions. Finally, a 3D Generative Adversarial Network (GAN)-based post-processing network is applied to enhance contextual consistency across adjacent slices, improving anatomical coherence in the reconstructed synthetic CT volumes.

The remainder of this paper is organized as follows. Section 2 reviews the state of the art in MR-to-CT translation and spine image synthesis. Section 3 describes the materials and methods, including the proposed three-stage framework and advanced validation strategies. Section 4 presents the experimental results and discusses the findings and their clinical implications. Finally, Section 5 summarizes the main conclusions and outlines future research directions.

\section{State of the art}
\label{sec:stateoftheart}

Medical image translation has attracted increasing research attention in recent years, with a rapidly expanding body of work focused on cross-modal synthesis \citep{mcnaughton2023machine}. Among the proposed approaches, deep learning methods have consistently outperformed conventional atlas, registration, and voxel-based techniques, establishing themselves as the dominant paradigm for the task \citep{dayarathna2024deep}. 

Early deep learning methods relied on Variational Autoencoders (VAEs) and fully convolutional neural networks (CNNs) for image synthesis \citep{cao2020auto, nie2016estimating}. However, limitations in image realism and structural fidelity motivated the widespread adoption of Generative Adversarial Networks (GAN), including paired and unpaired formulations such as Cycle-Consistent Generative Adversarial Networks (CycleGANs) \citep{goodfellow2014generative, gu2023cross, zhao2022egsde}. More recently, Transformer-based architectures and Denoising Diffusion Models have achieved state-of-the-art performance, surpassing earlier methods in image quality, anatomical consistency, and synthesis diversity \citep{vaswani2017attention, ho2020denoising}.

The following subsections review these methodological families, beginning with GAN-based approaches, followed by Transformer- and diffusion-based methods, and concluding with a focused review on MR-to-CT translation of spine images.

\subsection{\textit{GAN-based approaches}}

Generative Adversarial Networks (GANs) have become a central paradigm for medical image synthesis, relying on an adversarial objective between a generator and a discriminator to produce realistic outputs \citep{goodfellow2014generative}. 

These methods rapidly became a dominant approach for MR-to-CT synthesis, with several adaptations proposed to address challenges specific to medical imaging. For instance, \citep{nie2018medical} introduced a context-aware GAN, while \citep{armanious2020medgan} improved image quality and reduced blurring through the incorporation of additional non-adversarial losses. Other studies focused on improving anatomical fidelity during synthesis; for instance, \citep{emami2021sa} incorporated organ segmentation constraints to better preserve shape and spatial localization.

A major limitation of early GAN-based translation methods was their reliance on paired datasets, which are difficult to obtain. To overcome this issue, unpaired GAN variants were introduced, most notably CycleGANs \citep{zhu2017unpaired, gu2023cross, zhao2022egsde}. By incorporating a cycle-consistency loss, these models enforce reconstruction of the original image after forward and inverse domain translations, enabling training on unpaired MR and CT data while constraining unrealistic cross-modal mappings. This represented a major advance for MR-to-CT synthesis and led to successful applications across multiple anatomical regions, including the brain \citep{yang2020unsupervised, hiasa2018cross} and pelvis \citep{lei2019mri}. By combining adversarial and cycle-consistency objectives, these architectures improve cross-modal synthesis while maintaining anatomical plausibility. Nevertheless, GAN-based approaches remain limited by training instability and residual inconsistencies, which can compromise synthesis robustness and structural precision.
	 
\subsection{\textit{Transformers and diffusion models}}
\label{sec:transformer_and_diffusion_models}
Recent advances in medical image translation have increasingly leveraged transformer architectures and diffusion models, which have emerged as state-of-the-art alternatives to GAN-based approaches \citep{vaswani2017attention, ho2020denoising}. These methods are particularly effective at capturing both long-range dependencies and fine local details, enabling more coherent and anatomically consistent cross-modal synthesis.

Transformer-based methods introduce self-attention mechanisms that model relationships across the entire image, allowing the network to capture global contextual information. This characteristic is especially advantageous in MR-to-CT translation, where preserving large-scale anatomical relationships is critical. Several promising approaches have been proposed in this direction, including pure Transformer-based frameworks \citep{zhong2023multi} as well as hybrid architectures that combine attention mechanisms with adversarial learning \citep{zhao2023ct}.

Diffusion models constitute another major methodological advance in medical image synthesis. These models learn the reverse of a progressive noising process, enabling high-fidelity image generation through iterative denoising steps. In MR-to-CT translation, several unsupervised diffusion-based approaches have shown promising performance \citep{li2023zero, zhao2022egsde, luo2024target}. Other studies have explored hybrid formulations combining diffusion and adversarial objectives \citep{wang2022diffusion}.

In general, Transformer- and diffusion-based models have demonstrated superior synthesis quality compared with earlier GAN-based approaches, particularly in terms of anatomical coherence and image fidelity \citep{dhariwal2021diffusion}. However, these gains are often achieved at the cost of substantially higher computational complexity and memory requirements, which remain important practical limitations for high-resolution medical imaging tasks.

\subsection{\textit{MR-to-CT translation of spine images}}
\label{sec:MR-to-CT translation}
Several GAN- and CycleGAN-based approaches have been proposed for MR-to-CT translation of spine images, including the works of \citep{karthik2021three, kim2025translation, roberts2023imaging}. While these studies demonstrated the feasibility of spine image synthesis, important challenges remain, including limited dataset size, high variability in synthesized volumes, and restriction to specific spinal regions.

To address some of these limitations, more recent studies have explored diffusion-based methods. For example, \citep{chen2024contourdiff} proposed a diffusion model for sagittal lumbar spine MR-to-CT translation, highlighting the potential of iterative generative frameworks for improved structural fidelity.

Comprehensive evaluations of this task have also been reported, such as the review by \citep{graf2023denoising}, which compared several supervised and unsupervised architectures on spine datasets. Among the evaluated methods, SynDiff—an adversarial diffusion framework originally introduced by \citep{ozbey2023unsupervised}—stands out for its ability to leverage unpaired data while maintaining strong anatomical consistency. Notably, its translation performance was also evaluated on pelvic MR images containing the lumbar vertebrae and sacrum, further supporting its suitability for spine-related synthesis tasks.

However, challenges specific to full-spine imaging remain insufficiently addressed, particularly vertebral correspondence across unpaired datasets and slice-to-slice consistency in 3D volumes. These limitations motivated the use of SynDiff as the baseline architecture and directly informed the design of the proposed framework, which incorporates a vertebral matching system and a 3D consistency enhancement step to overcome the challenges of unpaired, slice-based 2D translation.

\section{Materials and methods}
\label{sec:materialmethods}

The following subsections describe the employed data and our proposed pipeline for unpaired MR to CT spine image translation.

\subsection{\textit{Dataset collection and preparation}}
\label{sec:MR_and_CT_datasets}
This study was conducted using publicly available datasets that collectively cover all spinal regions, including the cervical, thoracic, and lumbar spine.

Regarding MR data, datasets were preferentially selected to include sagittal T1-weighted acquisitions, as this plane is the clinical standard for global spine assessment and provides optimal visualization of vertebral structures.

For the cervical region, the Spine Generic Project and the Instrumented Spinal Cord T1/T2-weighted MRI Scans datasets were included \citep{cohen2021generic, sharafi2024instrumented}. For the lumbar region, the SPIDER dataset was selected \citep{van2023spider}. Publicly available MR scans covering the thoracic region or the full spine were considerably less common, as sagittal acquisitions spanning this area are less frequently available. Nevertheless, suitable cases were identified in the TotalSegmentator MRI Dataset \citep{wasserthal2025totalsegmentator}. This process yielded a total of 560 MR scans. 

\begin{figure*}[btp]
  \centering
  \includegraphics[width=1\textwidth]{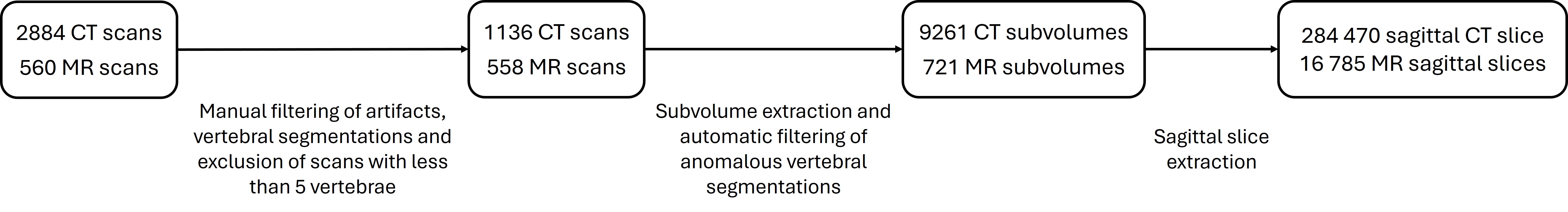}
  \caption{Data flow diagram illustrating the full pipeline from raw data to the final dataset construction. Starting from 2844 CT scans and 560 MR scans, vertebral segmentations are obtained via TotalSegmentator, followed by manual filtering and region-based subvolume extraction. Subvolumes are then automatically filtered based on the quality of their segmentation masks, discarding those with anomalous or incomplete vertebral segmentations. Since the downstream translation model operates on 2D, sagittal slices are extracted from the subvolumes. Finally, the resulting slices are split into training, validation and test sets.}
  \label{fig:data_flow}
\end{figure*}

For CT data, datasets were selected to provide analogous regional coverage across the same spinal sections. The cervical region was represented by the PDDCA, HaN-Seg, and HNSCC-3DCT-RT datasets \citep{raudaschl2017evaluation, podobnik2023han, bejarano2018head}. For the thoracic, lumbar, and multi-region scans, the Spine-Mets-CT-Seg, TotalSegmentator CT Dataset, and CTSpine1K datasets were incorporated \citep{spinemetsct2024, wasserthal2023totalsegmentator, deng2021ctspine1k}, producing a multisource CT dataset of 2884 scans.

After initial dataset identification, only scans containing at least five vertebrae, according to their vertebral segmentation, were retained. Some of these segmentations were already available in the original datasets, whereas the missing ones were generated using the TotalSegmentator tool~\citep{wasserthal2023totalsegmentator, wasserthal2025totalsegmentator}. 

In addition, cases presenting severe acquisition artifacts, such as low resolution, extreme intensity areas due to the presence of metal implants or excessive image noise (like quantum noise), were also excluded after visual inspection. Hence, the final imaging cohort comprised 558 MR scans and 1136 CT scans coming from 10 distinct open-access datasets. Figure \ref{fig:data_flow} summarizes the dataset generation process.

\subsubsection{\textit{Preprocessing}}
The first step of the preprocessing pipeline consisted of standardizing the imaging studies to reduce the intra-modality variability across MR and CT datasets. All scans were first reoriented to a LIP coordinate system and resampled to a voxel spacing of 1 mm x 1 mm x 2.5 mm. Intensity normalization was performed using min-max scaling. In addition, modality-specific preprocessing was applied: CT intensities were clipped to the range [-1000, 1000], while MR scans underwent bias field correction \citep{tustison2010n4itk}. 

\begin{table}
\centering
\footnotesize
\setlength{\tabcolsep}{3pt}
\renewcommand{\arraystretch}{1.15}
\resizebox{\columnwidth}{!}{%
\begin{tabular}{l l r r r r}
\toprule
\textbf{Split} & \textbf{Region} & \textbf{MR} & \textbf{CT} & \textbf{MR Slices} & \textbf{CT Slices} \\
\midrule
\multirow{5}{*}{Train}     
    & Cervical               & 198 & 340  & 4532 & 10303  \\
    & Cervical \& Thoracic   & 13  & 334  & 297  & 9984   \\
    & Thoracic               & 12  & 3028 & 270 & 82984  \\
    & Thoracic \& Lumbar     & 1   & 293  & 24   & 8280   \\
    & Lumbar                 & 261 & 2412 & 6196 & 85416  \\
\midrule
\multirow{5}{*}{Validation}
    & Cervical               & 60  & 107  & 1371 & 3228   \\
    & Cervical \& Thoracic   & 8   & 98   & 185  & 2964   \\
    & Thoracic               & 7   & 887 & 139  & 24298  \\
    & Thoracic \& Lumbar     & 0   & 81   & 0    & 2273   \\
    & Lumbar                 & 78  & 713  & 1860 & 25198  \\
\midrule
\multirow{5}{*}{Test}      
    & Cervical               & 33  & 51   & 739  & 1560   \\
    & Cervical \& Thoracic   & 6  & 55  & 137  & 1670   \\
    & Thoracic               & 3   & 470  & 65   & 12752 \\
    & Thoracic \& Lumbar     & 1   & 38   & 21   & 1078   \\
    & Lumbar                 & 40  & 354  & 949  & 12482  \\
\bottomrule
\end{tabular}%
}
\caption{Dataset split summary after sagittal slice extraction. MR and CT columns indicate the number of subvolumes per modality, while MR Slices and CT Slices reflect the total number of 2D sagittal slices extracted from the subvolumes. The statistics are further stratified by spinal region. A vertebral region is considered present if it contains at least 50\% of the possible vertebrae from that region (e.g 4 out of 7 cervical vertebrae).}
\label{tab:dataset_images_summary}
\end{table}

\begin{figure*}[hbtp]
  \centering
  \includegraphics[width=1\textwidth]{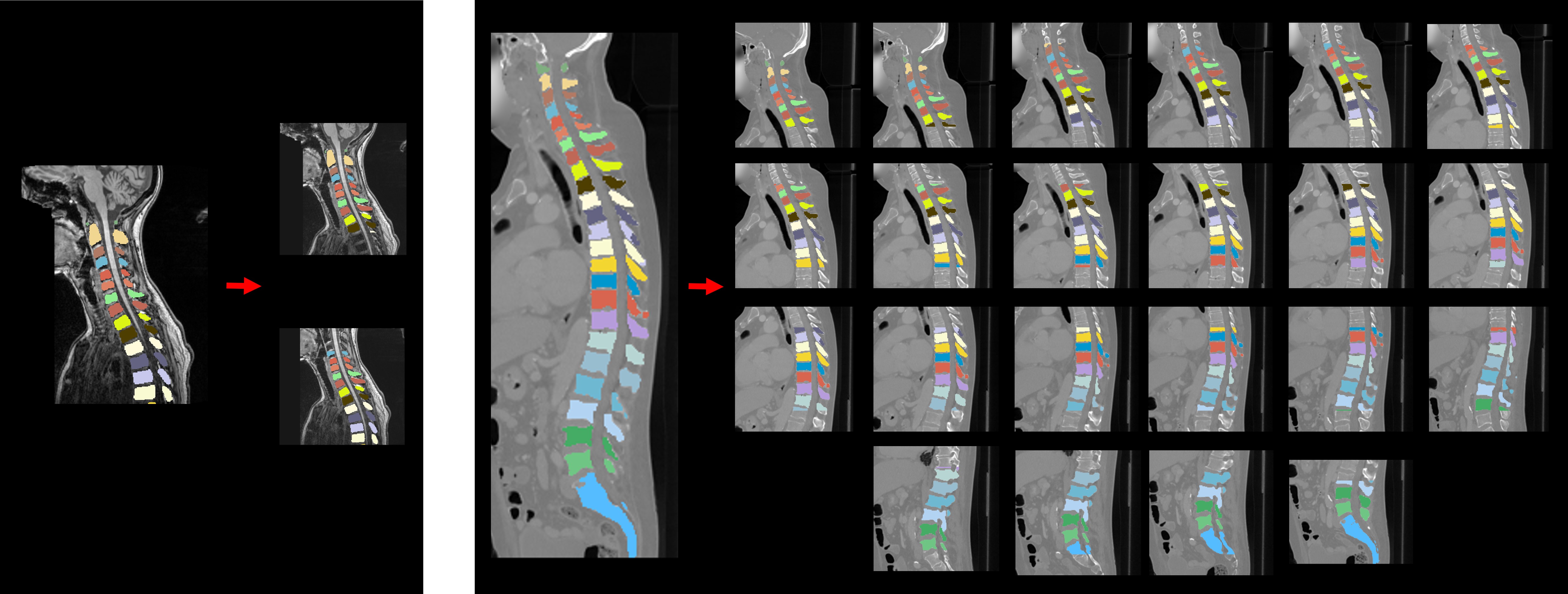}
  \caption{Example of MR (left) and CT (right) patches extraction after preprocessing. From each scan, multiple overlapping patches covering different spinal regions are extracted, as shown by the individual patch examples on the right of each scan. Unlike CT, where redundant patches from the same scan are allowed, MR patches are filtered to avoid patches with similar spine information, resulting in a significantly smaller number of extracted MR patches overall.}
  \label{fig:patches_extraction}
\end{figure*}

After this initial standardization, each scan was cropped in two stages using their spine segmentations. First, a general crop was applied to each scan to remove regions without spinal structures. After this first cropping, multiple 3D subvolumes were extracted using a sliding-window strategy, as shown in Figure \ref{fig:patches_extraction}. Specifically, the superior endplate of each vertebra was used as a reference, and 288x288xD subvolumes, where D corresponds to the original extent along the craneo-caudal axis, were extracted along the caudal direction. This patch size was selected empirically, since it provided enough anatomical coverage to include up to five vertebrae in any spinal region while avoiding excessively large empty areas. 

Since this strategy caused the inferior part of the scan to be excluded in most cases, a final 288x288xD subvolume was included in the extraction process, starting from the inferior end of the scan and extending in the superior direction. In this way we ensured complete coverage of the spinal column. 

In order to avoid data  redundancy, the sliding window was constrained so that extracted subvolumes from the same scan shared at most two complete vertebrae. This constraint was checked through the segmentation masks.

A second redundancy control criterion was also applied to intra-patient subvolumes by limiting the similarity between their spine segmentations to 90\%. This similarity was computed through the number of voxels per label in each subvolume, and was defined as:

\vspace{9pt}
\begin{equation}
    S(P_1, P_2) = 1 - \frac{\sum_{v \neq 0} \left| n_v^{P_1} - n_v^{P_2} \right|}{\sum_{v \neq 0} \max\left(n_v^{P_1}, n_v^{P_2}\right)}
    \label{eq:patch_similarity}
\end{equation}
\vspace{9pt}

Where $S(P_1, P_2)$ is the similarity score between subvolumes $P_1$ and $P_2$, $v$ denotes each vertebral label (excluding the background, $v \neq 0$), and 
$n_v^{P_i}$ is the number of voxels belonging to vertebra $v$ in patch $P_i$. A score close to 1 indicates that both subvolumes share a nearly identical vertebral volume distribution, and therefore cover the same spinal region, while a score close to zero indicates they cover different spinal regions.

Both redundancy filterings were only applied to MR subvolumes. No equivalent restriction was imposed on CT subvolumes, since redundancy in the CT pool was not detrimental for the subsequent pseudo-pair generation stage and retaining a larger number of CT candidates was beneficial for later vertebral matching.

Since the downstream translation model operates on 2D sagittal images, the final inputs to the network were the sagittal slices extracted from the resulting 3D subvolumes. Slices containing less than four vertebrae were discarded in order to exclude data with insufficient vertebral context before training. 

After this process, the dataset was divided into training, validation, and test sets with target proportions of 70\%, 20\%, and 10\%, respectively. To prevent data leakage, all subvolumes and corresponding slices from the same patient were assigned to the same dataset. In addition, the split was designed to preserve a representative distribution of vertebral region combinations across subsets. The preprocessing approach and a summary of the final dataset characteristics are shown in Figure \ref{fig:data_flow} and Table \ref{tab:dataset_images_summary}, respectively.

\subsection{\textit{Proposed approach}}
In this work, we propose a three-stage framework for MR-to-CT spine image translation. The first stage involves implementing a vertebral matching system, which processes unpaired MR and CT sagittal slices to generate pseudo-paired datasets by aligning anatomical structures. This step is crucial for addressing anatomical discrepancies and ensuring suitable inputs for the next stage. The second stage focuses on the core 2D image translation task, where we leverage the SynDiff neural network architecture (described in Section \ref{sec:MR-to-CT translation}). Finally, the third stage of the framework applies a post-processing network to enhance 3D contextual consistency across adjacent slices, ensuring improved anatomical coherence in the reconstructed synthetic CT volumes obtained by stacking the translated sagittal slices. This post-processing step is essential for refining the quality and structural accuracy of the synthetic CTs, and is implemented as a 3D GAN with a ResUNet generator and a PatchGAN discriminator.

In the following subsections, we provide detailed information on each stage of the framework, including the training process, model evaluation, and validation strategies.

\subsubsection{\textit{Vertebral matching  system}}
\label{sec:images_matching_system}
To enhance the performance of the MR-to-CT translation task, we introduce a vertebral matching system as the first stage of our framework. While SynDiff is designed to handle unpaired data, training with randomly selected MR and CT slices can lead to mismatches, especially in terms of spine region or vertebral alignment, which can negatively impact translation accuracy. The vertebral matching system mitigates this issue by ensuring that each MR slice is paired with a CT slice that has a similar spinal structure. 

As both MR and CT scans contain thousands of slices, making pixel-wise comparison across all possible pairs is computationally prohibitive. To address this, we first compute a simple shape descriptor for each MR and CT sagittal slice segmentation, derived from the relative proportion of pixels occupied by each vertebrae (C1, C2... sacrum) in the complete segmentation mask. This descriptor is represented as a 1x25 array, and the cosine distance between the descriptors extracted from MR and CT slices are used to identify the most relevant MR-CT slice match candidates (two pair of images with the same vertebrae and with a similar size have closer descriptors).

While this approach helps identify slices with similar vertebral proportions, it does not guarantee identical spine shapes or vertebral tissue overlap, as the pixel distribution may still differ and the scans may be misaligned. Therefore, the descriptor is used only for an initial filtering, followed by a fixed registration, inspired by the work of \citet{graf2023denoising}, and a second comparison using a weighted Dice Similarity Score (DSC), which accounts for vertebral size and penalizes the presence of additional vertebrae. This weighted DSC is defined as:

\vspace{10pt}
\begin{equation}
    \text{DSC}_w = \frac{\sum_{c \neq 0} \text{DSC}_c \cdot w_c}{\sum_{c \neq 0} w_c}, \quad
    \text{DSC}_c = \frac{2|A_c \cap B_c|}{|A_c| + |B_c|}
    \label{eq:weighted_DSC}
\end{equation}
\vspace{8pt}

Where $A_c$ and $B_c$ are the binary masks of class $c$ in each segmentation and and $w_c = |A_c| + |B_c|$ is defined as the total number of voxels of class $c$ across both masks. The background class ($c=0$) is excluded from the metric computation.

This two comparisons ensure that not only a shape correspondence between images is established, but also tissue localization and overlap. 

Regarding the number of comparisons executed for each slice, the fifty most similar candidates according to the descriptor are selected, restricting the search to those differing by at most one vertebra. For each candidate, the CT spine segmentation is aligned with the MR spine through a rigid registration based on vertebrae landmarks. These landmarks correspond to the vertebral body and spinous processes centroids in both scans, and the registration applied is a 2D Euler transformation \citep{lowekamp2013design}.

\begin{figure*}[ht]
  \centering
  \includegraphics[width=0.9\textwidth]{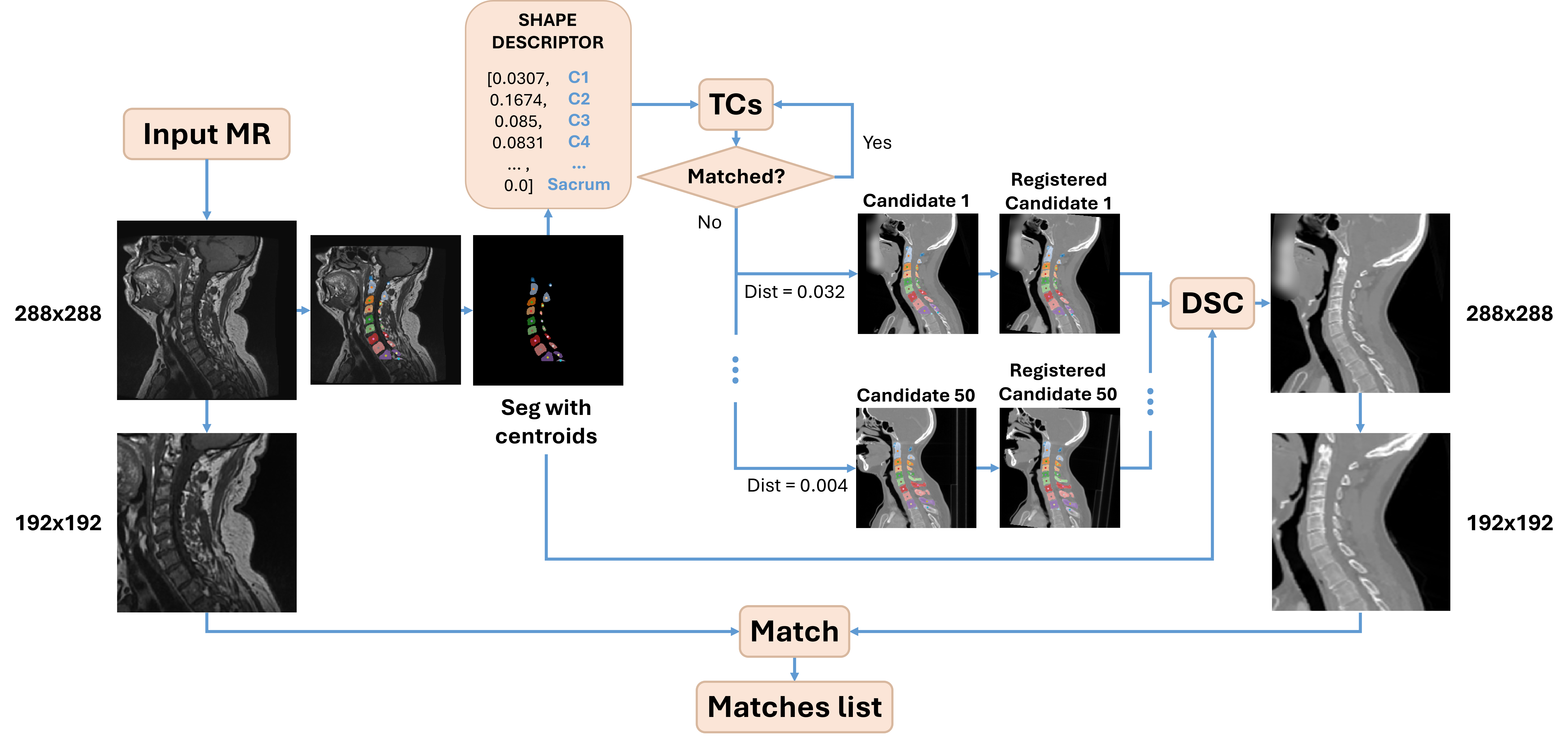}
  \caption{Vertebral matching system overview. Figure illustrates how every MR sagittal slice is paired with a slice of the CT corresponding set, trying to maximize the spine similarity and overlap between each pair through a rigid registration process. Once paired, both slices are cropped to avoid empty pairings resulting from excessive rotations or shifts from rigid registration.}
  \label{fig:matching_system}
\end{figure*}

Once aligned, both segmentations are compared through the weighted DSC, and the final match is then chosen as the one with the highest DSC among the fifty candidates. Candidates are also discarded if the DSC obtained is equal to zero. After a CT slice is matched, it is removed from the possible candidates for the following MR slices, and the rigid transform applied to the segmentation is also applied to the corresponding CT image. Since this transformation may introduce empty regions due to shifts or rotations, the resulting image is cropped to the central 192x192 region. An overview of the proposed matching approach is shown in Figure \ref{fig:matching_system}.

\begin{figure*}[h]
  \centering
  \includegraphics[width=0.7\textwidth]{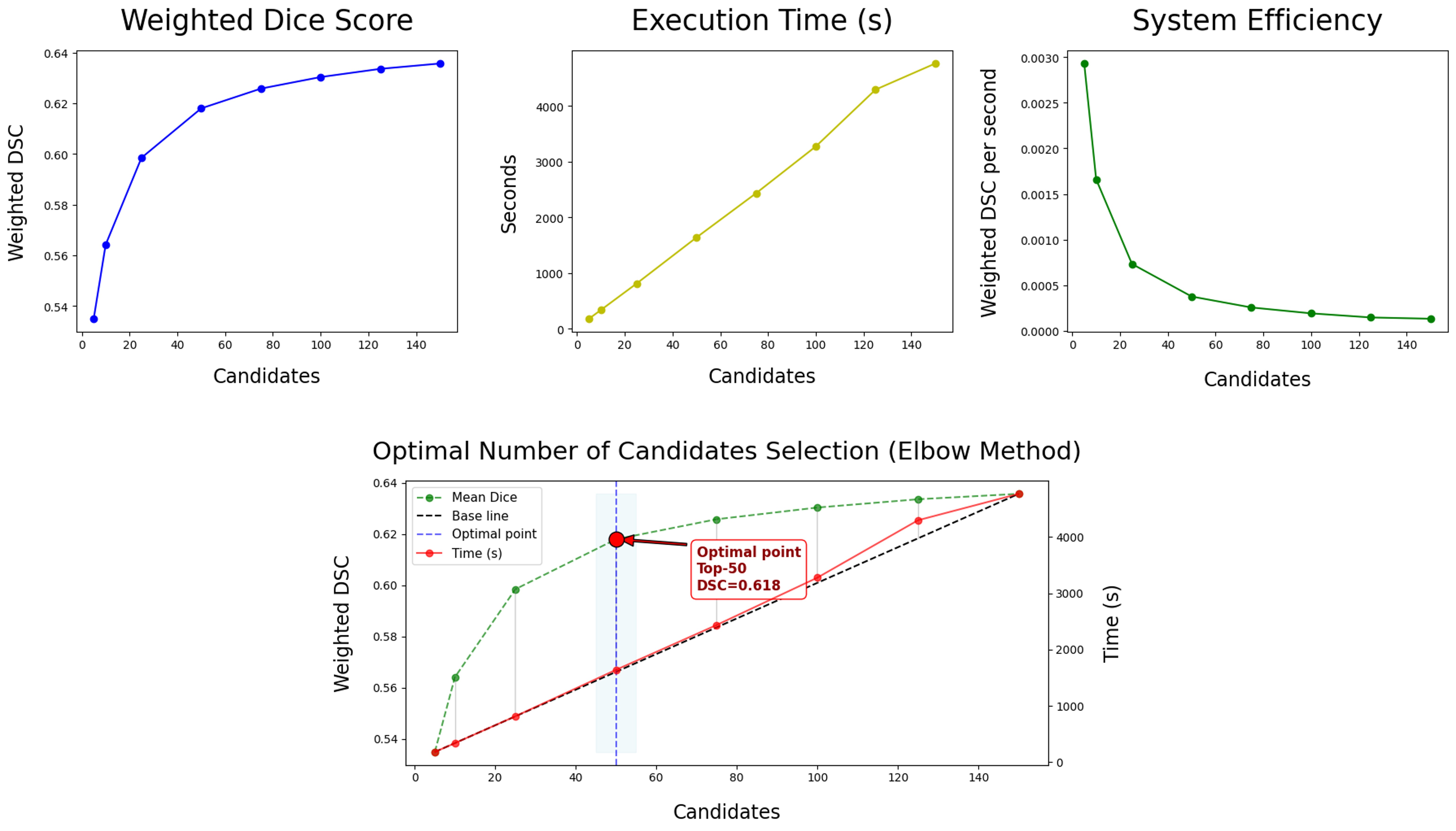}
  \caption{Vertebral matching system empirical evaluation on the test dataset. The top row shows the obtained weighted DSC, execution time and matching efficiency (DSC increase per second of execution) as a function of the number of evaluated candidates. The bottom plot applies the elbow method to find the optimal trade-off between patch spinal similarity (DSC) and computational cost, identifying 50 candidates as the optimal selection.}
  \label{fig:matching_sytem_plots}
\end{figure*}

Limiting the candidates to fifty strikes a balance between computational efficiency and matching accuracy, determined through empirical testing on the test dataset. As shown in Figure \ref{fig:matching_sytem_plots}, increasing the number of candidates improves accuracy but disproportionately increases computational time. The elbow method was applied to the efficiency curve, identifying fifty candidates as the optimal trade-off point.

However, if no match is found within the first fifty candidates because there are no slices left with the same vertebrae, the system expands the candidate pool and also allows for a greater vertebral difference.

Following this system, the MR train, validation and test sets were matched with the most similar slices belonging to the corresponding CT set. The matching results were as shown in Table \ref{tab:registered_matching_results}.

\begin{table}[h]
\centering
\small
\begin{tabular}{lccccc}
\toprule
\textbf{Dataset} & \makecell{\textbf{MR-CT} \\ \textbf{pseudo-pairs}} & \makecell{\textbf{Mean} \\ \textbf{DSC}} & \multicolumn{3}{c}{\textbf{Differing vertebrae}} \\ 
 & & & \textbf{0} & \textbf{1} & \textbf{$>1$} \\
\midrule
Train & 11319 & 0.6374 & 6459 & 4855 & 5 \\
Validation & 3555 & 0.6340 & 1825 & 1728 & 2 \\
Test & 1911 & 0.6180 & 975 & 928 & 8 \\
\bottomrule
\end{tabular}
\caption{Vertebral matching system results for the different datasets. The mean DSC value represents the average overlap between the MR and CT spines in the generated pseudo-pairs. The differing vertebrae indicates the number of vertebral mismatches between the sagittal slices of every generated pseudo-pair, with 0 meaning both slices have the same vertebrae, 1 meaning there is at least one vertebra that is not present in both slices and >1 meaning there is more than 1 vertebrae that is not present in both slices.}
\label{tab:registered_matching_results}
\end{table}

\subsubsection{\textit{MR to CT translation using SynDiff}}
\label{sec:SynDiff_training}

Following the vertebral matching system introduced earlier, we proceeded to train the SynDiff architecture.

SynDiff consists of two main components: a non-diffusive module and a diffusive module. The non-diffusive module generates an approximation of the target modality based on features extracted from the input MR slice. The diffusive module then takes the synthetic approximation and the noisy MR slice, learning to progressively denoise and refine the image to match the target CT modality. Both modules are structured as CycleGANs, ensuring that any transformations applied maintain anatomical consistency between the input and the generated target image.

For training, we performed an ablation study to evaluate the impact of incorporating the vertebral matching system. Specifically, we compared the performance of two setups: one using MR-CT sagittal slice pairs from the same spinal region (cervical, thoracic or lumbar), without any vertebrae matching or alignment, and the other one using the vertebral matching system to create pseudo-pairs based on vertebral structures similarity. For both approaches, the same MR sagittal slices from the extracted 3D subvolumes described in \ref{sec:MR_and_CT_datasets} were used.

Both setups were trained until stability was observed in terms of spine-related metrics. Using the original SynDiff architecture and based on the original training hyperparameters, the input sagittal slices for translation were padded from 192x192 to a size of 256x256 to match the input size used in the original implementation. Other training parameters included a batch size of one, a learning rate of $1 \times 10^{-4}$, two residual blocks for the non-diffusive generator, 1000 direct diffusion steps, and four denoising steps for the diffusive generator.

\subsection{\textit{3D contextual consistency enhancement}}
The final stage of the proposed pipeline focuses on improving the 3D reconstruction quality of the generated synthetic CT volumes. Since the MR-to-CT translation network operates in a slice-wise 2D manner, the independently generated sagittal slices may present local discontinuities when reassembled into a full 3D volume. To enhance inter-slice coherence and improve anatomical continuity across adjacent vertebral structures, a dedicated post-processing network was introduced. This refinement strategy aims to preserve the high in-slice image quality achieved by SynDiff while improving volumetric consistency, without requiring modifications to the core translation architecture.

\begin{figure}
  \centering
  \includegraphics[width=0.6\columnwidth]{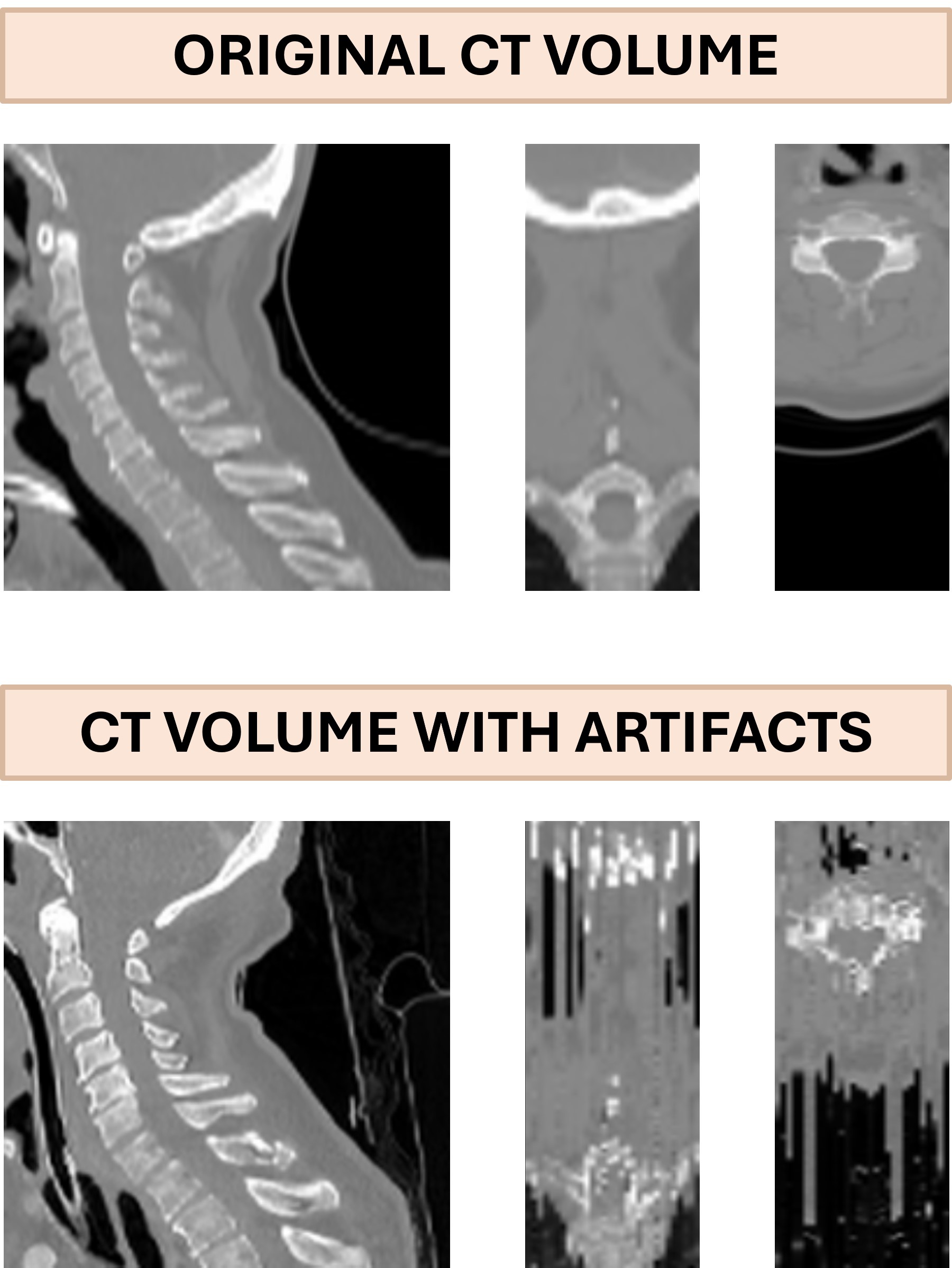}
  \caption{Example of the simulated discontinuity artifacts generated through the vertebral matching system a on real CT volume.}
  \label{fig:sim_artifacts_example}
\end{figure}

To train this refinement module in a supervised manner, a representative subset corresponding to approximately 10\% of the CT train subvolumes described in Figure~\ref{fig:data_flow} was selected as training data. For each of this 670 real CT subvolumes, a corresponding pseudo-CT simulating the discontinuity artifacts observed in the stacked synthetic CT volumes, was generated by adapting the vertebral matching system described in Section~\ref{sec:images_matching_system} to operate exclusively within the CT domain. These new volumes were generated by matching each sagittal slice from the input CT volumes to the most anatomically similar slice extracted from the remaining CT scans, preserving local vertebral appearance while intentionally disrupting inter-slice continuity across the reconstructed volume as illustrated in Figure~\ref{fig:sim_artifacts_example}. The resulting pseudo-CT volumes were used as network inputs, whereas the original anatomically coherent CT volumes served as ground-truth targets for supervised learning.

For testing, the 83 synthetic CT volumes generated from reassembling the 2D translated MR sagittal slices of the test split described in Figure~\ref{fig:data_flow} were used.

\begin{figure*}[h]
  \centering
  \includegraphics[width=1\textwidth]{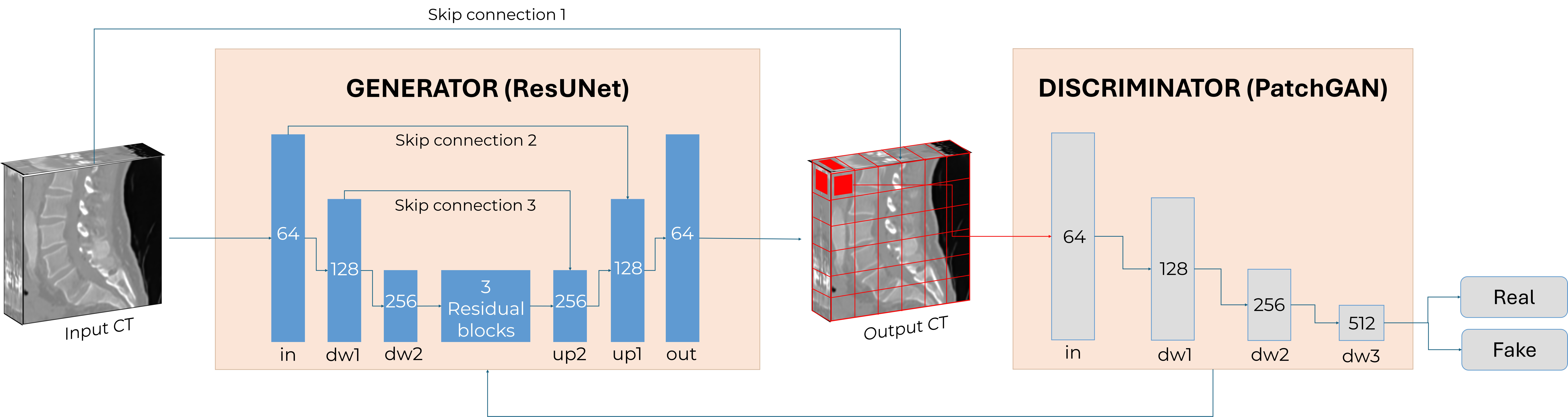}
  \caption{Overview of the post-processing net architecture. The ResUNet generator takes a synthetic CT as input and produces a corrected CT output, trying to increase its volumetric consistency and coherence. Residual skip connections concatenate encoder and decoder feature maps, enabling multi-scale feature fusion and improved anatomical detail preservation. The PatchGAN discriminator classifies local regions of the output as real or fake to provide adversarial supervision.}
  \label{fig:post_processing_net_architecture}
\end{figure*}

Regarding the architecture of the post-processing network, it was implemented as a supervised 3D GAN, with a residual encoder–decoder generator and a volumetric PatchGAN discriminator (Figure~\ref{fig:post_processing_net_architecture}).

The generator was designed as a fully 3D convolutional encoder–residual–decoder architecture. The encoder progressively reduced spatial resolution through strided 3D convolutions followed by instance normalization and ReLU activations. At the bottleneck, three residual blocks were applied to learn non-linear volumetric representations of inter-slice inconsistencies. Spatial resolution was then restored through two decoder blocks based on trilinear upsampling followed by 3D convolutions, instance normalization, and ReLU activations. To preserve fine anatomical details, skip connections linked the encoder and decoder at matching resolutions. These skip connections were further refined through additional residual units, enabling selective preservation of anatomically consistent regions while correcting local discontinuities. Additionally, a direct skip connection from the input to the output was incorporated to preserve global structural information. The final output synthetic CT volume was generated through a 3D convolution, with output values constrained to [0,1] using a clipping operation.

The discriminator followed a 3D PatchGAN design, providing adversarial feedback at the level of overlapping volumetric patches rather than the full volume. It consisted of consecutive 3D convolutional blocks with instance normalization and LeakyReLU activations, progressively increasing the number of feature channels while reducing spatial resolution. This patch-based formulation encouraged the correction of local high-frequency discontinuities and improved robustness to regional artifacts.

During training, the generator was optimized using a weighted combination of adversarial, L1, SSIM, perceptual and inter-slice consistency losses. The interslice consistency term was defined as the L1 distance between consecutive sagittal slices, encouraging consistent transitions.  The perceptual loss was computed from feature representations extracted using a pretrained 3D ResNet-18 network. The total generator objective is defined in Equation~\ref{eq:total_loss}, where the loss weights were selected through Optuna-based hyperparameter optimization to maximize refinement performance \citep{akiba2019optuna}.

\vspace{9pt}
\begin{align}
g_{\text{total\_loss}} &= 0.2105 \cdot \text{adv\_loss} 
                         + 0.1145 \cdot \text{l1\_loss} \nonumber\\
                         &\quad
                         + 0.1 \cdot \text{inter-slice\_consistency\_loss} \nonumber\\
                         &\quad
                         + 0.2562 \cdot \text{ssim\_loss} \nonumber\\
                         &\quad 
                         + 0.2629 \cdot \text{perc\_loss}
\label{eq:total_loss}
\end{align}
\vspace{9pt}

Both networks were trained with a batch size of 1 and learning rates were set to $4 \times 10^{-4}$ for the generator and $8 \times 10^{-5}$ for the discriminator.

\subsection{\textit{Validation methods}}
To determine the quality of the proposed approach beyond qualitative assessment, we computed probability distribution-based metrics, task-specific metrics and atlas-based metrics from full images and specific vertebrae regions. 

\subsubsection{\textit{Distribution-based metrics}}
\label{subsec:distribution-base-metrics}
Distribution-level similarity between synthetic and real CT images was evaluated using Maximum Mean Discrepancy (MMD) and Fréchet Inception Distance (FID) \citep{valdebenito2023exploration}. 

MMD was computed on image intensity distributions using an RBF kernel. Although the same kernel was used across comparisons, raw MMD values are not directly comparable across different distributions, as they depend on the data statistics and they do not have a fixed scale. In other words, it is not possible to determine whether an obtained MMD value is good or bad, as there are no fixed reference values defined. Therefore, we used the MMD between the real MR and real CT distributions, which are very different, as a high discrepancy value. A relative MMD (rMMD) was then defined as the ratio between the MMD between the synthetic CT and real CT distributions and the MMD between the real MR and real CT distributions.

FID was computed as a complementary feature-based realism metric using Inception-v3 embeddings. Given that the network is pretrained on natural images, FID was used only as a relative comparison measure. 

\subsubsection{\textit{Task-specific metrics}}
Task-specific evaluation was performed to assess the anatomical utility of the generated synthetic CT images through an auxiliary vertebral segmentation task.

Because synthetic CT in spine surgery is primarily intended to improve vertebral delineation for applications such as implant planning and pedicle screw placement, we quantified how accurately vertebral structures could be recovered from the generated images. For this purpose, Dice Similarity Coefficient (DSC) was computed between the vertebral segmentation obtained from the input MR image and the segmentation extracted from the corresponding synthetic CT, assuming preservation of vertebral spatial correspondence after translation.

Volumetric and slice level evaluation was performed using TotalSegmentator, which additionally provides an indirect measure of 3D anatomical coherence due to its reliance on spatial context.

\subsubsection{\textit{Atlas-based metrics}}
To further assess vertebral anatomical fidelity, atlas-based intensity metrics were computed through the reconstructed synthetic CT vertebrae.

Using the CT training set described in Section \ref{sec:MR_and_CT_datasets}, vertebra-specific reference intensity distributions were built by extracting Hounsfield Unit (HU) values from real CT images within the corresponding TotalSegmentator vertebral masks, excluding partially visible vertebrae at image borders. These reference distributions were then compared with those obtained from the synthetic CT reconstructions.

Comparison was performed using Earth Mover’s Distance (EMD) and Jensen-Shannon Divergence (JSD), providing complementary measures of distribution agreement in terms of global intensity shifts and histogram shape similarity, respectively.

\section{Results and discussion}

\begin{figure*}
  \centering
  \includegraphics[width=1 \textwidth]{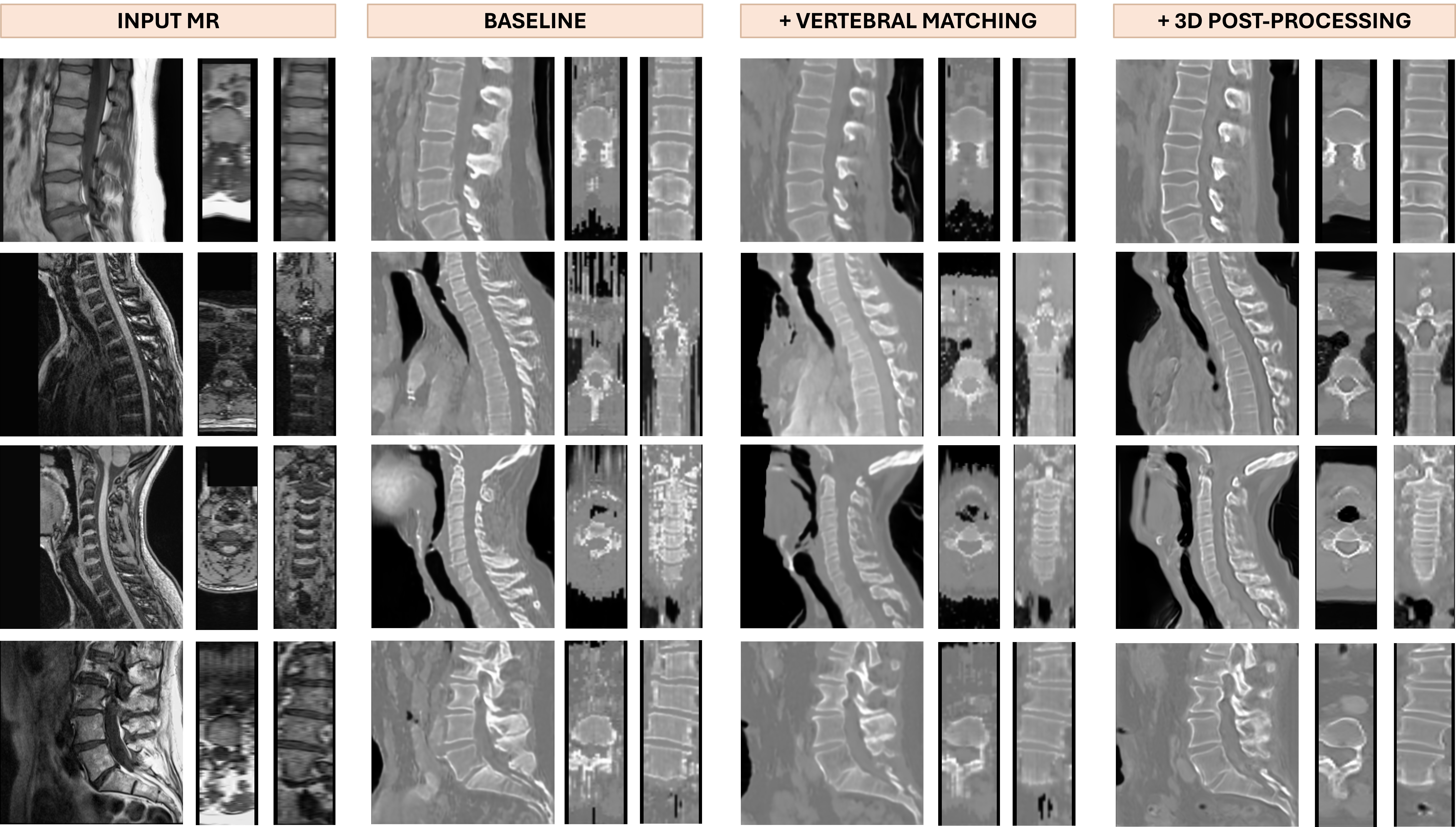}
  \caption{Progressive qualitative impact of each pipeline stage. Rows show representative and challenging examples across different spinal regions.}
  \label{fig:qualitative_results}
\end{figure*}

\begin{figure}
  \centering
  \includegraphics[width=0.9\columnwidth]{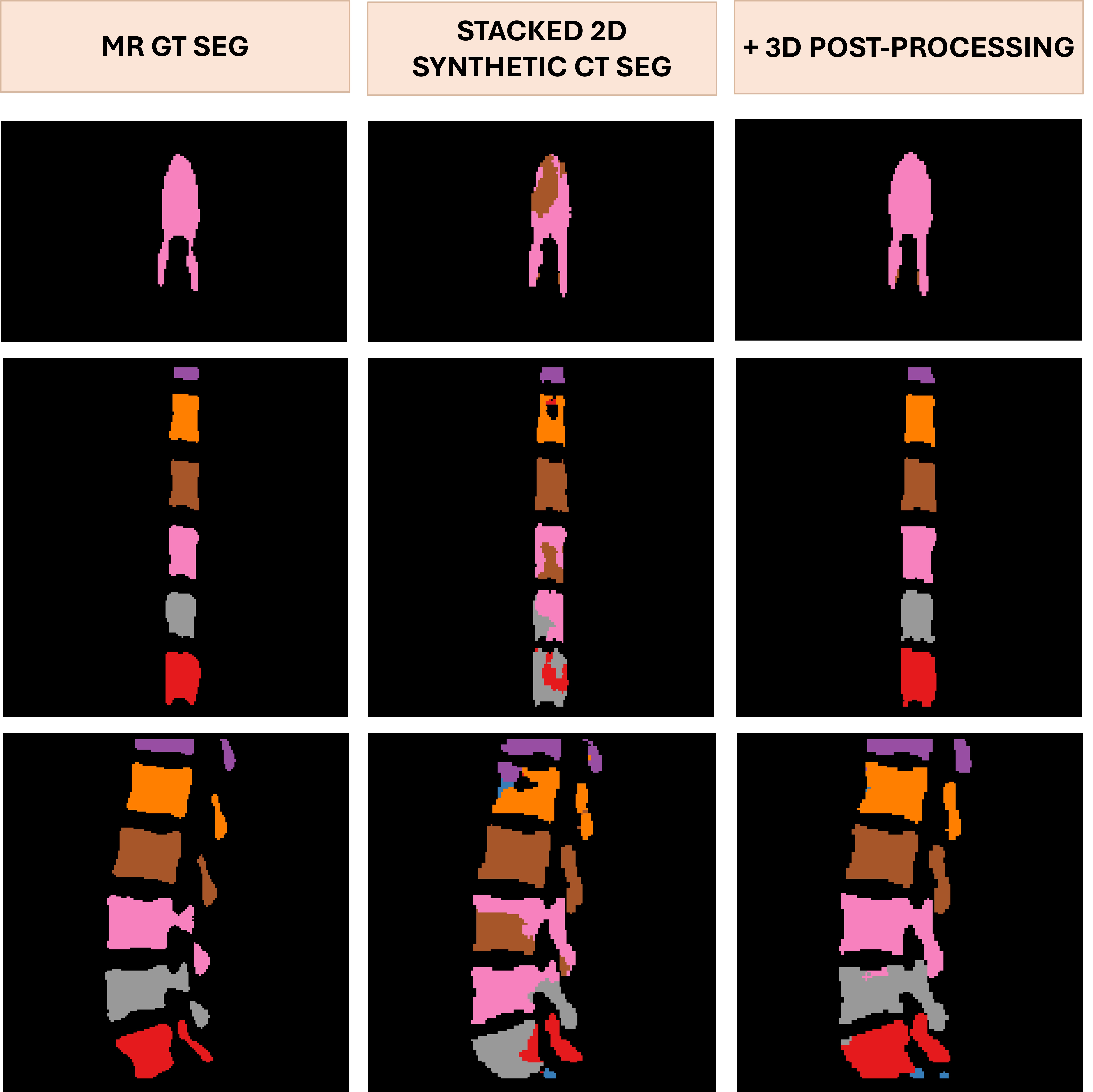}
  \caption{Volumetric vertebral coherence before and after 3D post-processing, seen through the vertebral segmentations in the different anatomical planes. The left column shows the MR ground truth segmentation, the middle column the synthetic CT segmentation obtained with the 2D translation model and the vertebral matching system approach and the right column shows the result after applying the 3D post-processing network.}
  \label{fig:volumetric_coherence}
\end{figure}

All experiments were evaluated on the independent test split introduced in Section~\ref{sec:MR_and_CT_datasets}, including 1911 sagittal MR slices from 83 3D subvolumes. Table~\ref{tab:incremental_ablation} summarizes the quantitative results of the proposed pipeline and Figure~\ref{fig:qualitative_results} reflects the visual results for different cases across the test set. Figure~\ref{fig:volumetric_coherence} shows the results related to volumetric consistency through vertebral segmentations of the different anatomical planes. 

In general, introducing the vertebral matching system prior to training improves the quantitative results over the baseline. This improvement is consistently reflected in a lower vertebra-wise FID and higher mean binary and multilabel 3D Dice scores, indicating better anatomical alignment and vertebral tissue quality. This confirms that anatomically coherent pseudo-pairs provide a stronger supervision signal than the unpaired setting.

Regarding the global alignment with the CT domain, the vertebral matching system approach reduces the rMMD compared to the baseline ([0.135] vs [0.053]), suggesting a closer match to the CT intensity distribution. However, this does not translate into an improved perceptual image quality, as measured by the Full FID, which increases slightly from 77.84 to 80.88. Given the relatively small magnitude of this difference, both approaches can be considered comparable in terms of global image quality.

Reconstructing the 3D volumes through the stacking of sagittal slices also benefits from the vertebral matching system, as shown by the 3D dice score, leading to more anatomically coherent reconstructed volumes when compared to the baseline approach. 

Regarding the qualitative results, it is further confirmed that both the baseline and vertebral matching system approaches learned an acceptable representation of the CT domain, with a clear delineation for the vertebral bodies. This can be appreciated especially in the lumbar and thoracic regions which are presented in row number one and row number two in shown in Figure~\ref{fig:qualitative_results}.

However, the baseline is less robust in regions with high variability across patients, making the translation of structures like the upper cervical vertebrae particularly challenging. More specifically, the C1 vertebra is sometimes merged with the C2 vertebra, leading to shape mismatches when compared to the input MR.

In addition, there are regions where both models struggle, especially the sacrum, which is sometimes merged with the L5 vertebra in certain volumes, resulting in a smoother transition when compared to the rest of the lumbar vertebrae. This may be again caused by the high variability of this region and because of its different anatomical characteristics when compared to the rest of the spine. An example of the baseline difficulties with the cervical region and of both models with the sacrum are shown in row number three and number four of Figure~\ref{fig:qualitative_results}.

Nevertheless, the regions which show the biggest anatomical differences across approaches are the spinous processes. While the baseline approach is more susceptible to produce bigger spinous processes and merge them across the spine, the pseudo-matching approach is able to produce more distinguishable and better delineated structures. This can be observed not only in sagittal slices, but also in the axial and coronal views, where spinous processes have more regular shapes.

\begin{table*}
\centering
\setlength{\tabcolsep}{3.5pt}
\begin{tabular}{l c c c c c c c}
\toprule
& \multicolumn{2}{c}{Global realism} & \multicolumn{5}{c}{Vertebral fidelity} \\
\midrule
\textbf{Method} & \textbf{rMMD ↓} & \textbf{Full FID ↓} & \textbf{TS 3D DSC (bin.)↑} & \textbf{TS 3D DSC ↑} & \textbf{Vertebral FID ↓} & \textbf{EMD ↓} & \textbf{JSD ↓} \\
SynDiff baseline & 0.135 & 77.84 & 0.666±0.07 & 0.435±0.17 & 41.57 & 111.2±61.1 & 0.308±0.13 \\
+ pseudo-matching & 0.053 & 80.88 & 0.739±0.04 & 0.564±0.15 & 33.64 & 85.7±44 & 0.267±0.13 \\
+ 3D post-processing & 0.074 & 87.41 & 0.732±0.04 & 0.575±0.133 & 50.37 & 76.8±63.2 & 0.213±0.14 \\
\bottomrule
\end{tabular}
\caption{Incremental contribution of each pipeline stage. Each row adds a component on top of the previous one. Global realism metrics evaluate the overall image quality, while vertebral fidelity metrics asses the anatomical consistency of the generated CT. TS 3D DSC reports the multiclass Dice per vertebra, while TS 3D DSC (bin.) treats all vertebrae as a single foreground class. Values are reported in a mean $\pm$ and standard deviation (when available) format computed over the 83 volumes of the test set. $\uparrow$ indicates higher is better, $\downarrow$ lower is better.}
\label{tab:incremental_ablation}
\end{table*}

Finally, the 3D post-processing network visually reduces the discontinuity artifacts observed in the axial and coronal planes of the 2D approaches. Vertebrae contours in all three anatomical planes become better defined, improving volumetric consistency, as reflected by the increase in TotalSegmentator 3D DSC from 0.564 to 0.575. However volumes also get smoother, modifying the intensity of the generated vertebrae and surrounding tissues, leading to the degradation of the realism metrics, with a slight increase in the full-image FID ([80.88] vs [87.41]) and the vertebral FID (33.64 vs 50.37). Although MMD also suffers a slight decline, its value is remains close to zero, meaning that the post-processed volumes are still aligned with the CT domain intensity distribution.

The effect of the post-processing network on volumetric anatomical coherence is further highlighted in Figure~\ref{fig:volumetric_coherence}. While the reconstructed 3D volumes obtained from the best 2D stage already preserved the overall vertebral arrangement and delineation, local discontinuities and vertebrae mismatch remained visible after slice stacking, mainly due to intensity discrepancies between predicted and real vertebrae. The proposed volumetric refinement reduced these discontinuities and produced smoother, anatomically more coherent vertebral reconstructions.

This effect is particularly evident in the 3D overlap renderings, where the refined synthetic CT segmentations show improved correspondence with the MR vertebral segmentations. These observations are consistent with the quantitative improvement observed in the volumetric TotalSegmentator DSC.

\begin{figure*}
  \centering
  \includegraphics[width=1 \textwidth]{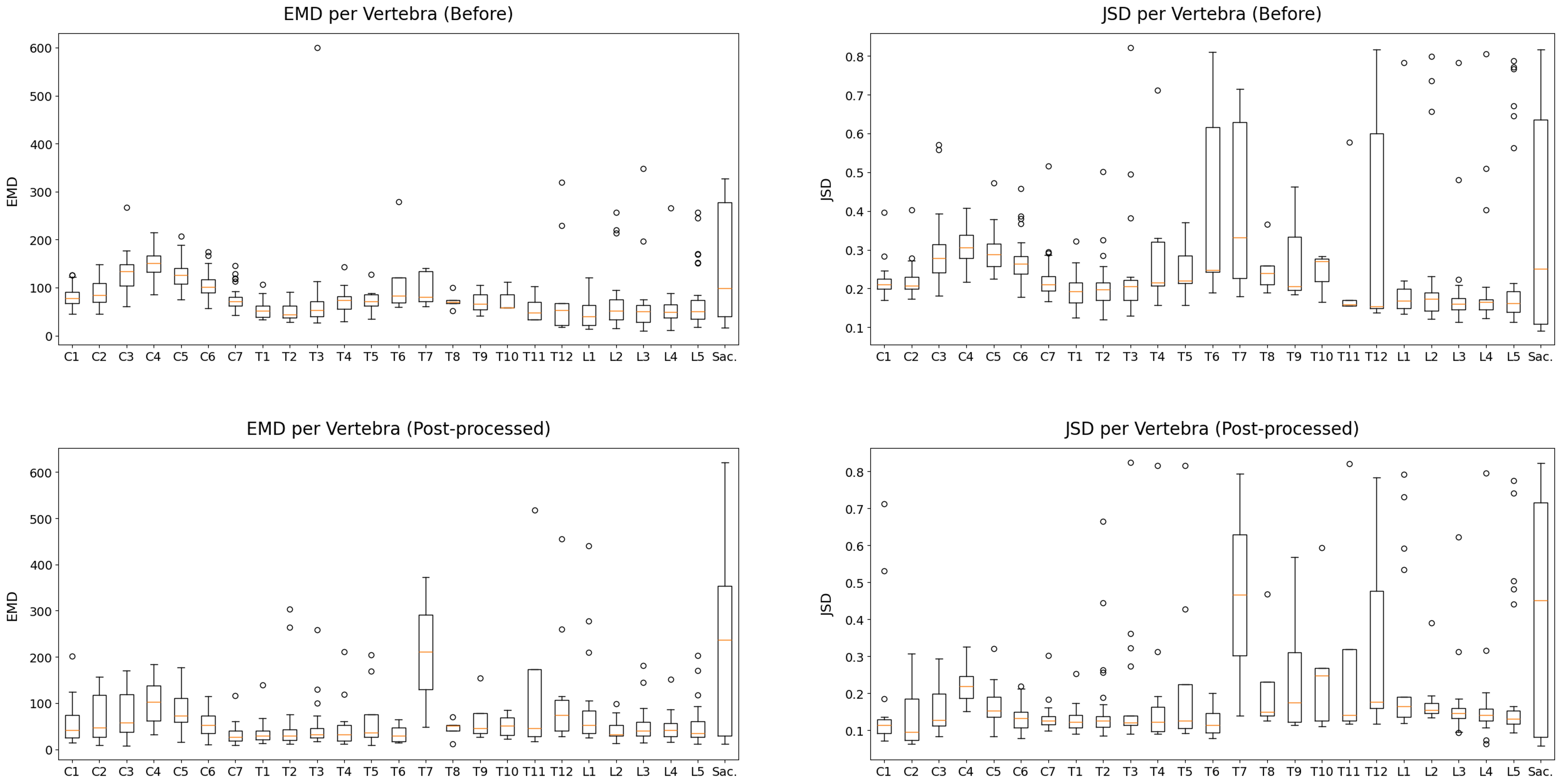}
  \caption{Vertebra-wise intensity fidelity before and after 3D refinement.}
  \label{fig:emd_and_jsd_box_plots}
\end{figure*}

To further analyze vertebral-level fidelity, Figure~\ref{fig:emd_and_jsd_box_plots} reports vertebra-wise EMD and JSD distributions before and after 3D refinement. The results reveal a heterogeneous behavior across vertebral regions for both the 2D MR-to-CT translation model and the post-processing network.

In the cervical region, high EMD and moderate JSD values are appreciated prior to 3D refinement. However, applying the post-processing network translates into a partial improvement. While the median values are reduced, the interquartile ranges are notably increased, indicating a higher resembling of the post-processed cervical vertebrae with the real ones but with a higher variability.

For the upper thoracic region (T1-T6), similar results can be appreciated. Prior to the refinement, the median values and interquartile ranges for both EMD and JSD are lower than for the cervical region. Still, after post-processing, there is a significant improvement in both EMD and JSD. This can be clearly appreciated for T6. 

In contrast, the lower thoracic region (T7-T12), exhibits significantly higher median values and interquartile ranges. In addition, 3D refinement is not able to reduce these values, increasing some of them and highlighting T7 as a challenging vertebra. These difficulties may be attributed to its lower anatomical representation in the training cohort.

As for the lumbar region, values remained similar before and after post-processing, with the best EMD and JSD values across the spine. Still, a slight degradation is observed at L1.

Finally, the sacrum remained the most challenging region, showing a significantly higher median and larger interquartile ranges for both EMD and JSD. These values further deteriorate after post-processing.

This behavior is consistent with the qualitative findings, where partial shape distortions and local smoothing persisted even after volumetric refinement. Nevertheless, the overall distribution alignment remained within an acceptable range for most vertebral levels, supporting the anatomical fidelity of the proposed framework.

Despite the overall improvements, several limitations remain. The sacrum and posterior vertebral elements continue to be the most failure-prone structures, likely due to their higher anatomical variability and comparatively lower representation in the available datasets. Similarly, thoracic performance remained less stable than cervical and lumbar regions, which may be explained by the scarcity of publicly available full-thoracic sagittal MR data. In addition, although the post-processing network improved volumetric coherence, it introduced mild smoothing in non-bone tissues, which negatively affected global full-image realism metrics. 

Finally, the lack of paired MR-CT ground truth prevents the use of voxel-wise similarity metrics and limits the evaluation to distributional and downstream task-based validation. Likewise, FID should be interpreted cautiously, as its natural-image feature extractor may not fully capture the anatomical characteristics of medical images.

\section{Conclusions}
\label{sec:conclusions}
In this work, we proposed a fully automatic pipeline for the translation of T1-weighted MR scans to CT scans. The method relies on a pseudo-pairing system which matches and aligns unpaired MR-CT sagittal slices with a similar vertebral anatomy through their spine segmentations, followed by a diffusive adversarial training with the SynDiff network. This pipeline was evaluated using a 2D approach followed by a post-processing network to better address spatial coherence in the predicted volumes.

Our experiments demonstrate that the 2D approach benefits significantly from the proposed vertebral matching system. This system enables more consistent spine alignment between modalities, resulting in improved structural fidelity in the synthetic CT scans, particularly in the vertebral regions. Although overall image quality metrics are slightly worse, this study was primarily oriented toward producing synthetic CT scans to support preoperative surgical planning, where anatomical accuracy in key regions is critical, with less focus on the translation of the rest of the structures.

From a volumetric perspective, the proposed 2D approach struggled to maintain anatomical coherence across axial and coronal views. Although precise and validated volumetric segmentation algorithms were able to achieve an acceptable 3D segmentation, qualitative results showed the presence of discontinuity artifacts across slices. 

The post-processing net, which aimed to solve this issue, successfully eliminated the artifacts and achieved a slight improvement in the vertebral anatomical fidelity across all planes. However, it also had a negative impact on the quality of the 2D sagittal slices, reducing the accuracy of vertebral tissue intensities and overall image realism.

This trade-off suggests that future work should explore refinement strategies that preserve 2D intensity fidelity and realism while improving vertebral continuity. 

Potential directions include the development of fully 3D translation networks that can better capture anatomical continuity across slices, more sophisticated post-processing techniques that reduce discontinuity artifacts while preserving fine-grained details, and the integration of additional anatomical priors or multi-modal constraints to further enhance structural fidelity. Moreover, exploring hybrid approaches that combine 2D efficiency with 3D spatial awareness may provide a practical balance between computational cost and volumetric accuracy.

Although the clinical adoption of synthetic CT generation tools may still face important regulatory and validation challenges, particularly in surgical applications where reliability and safety are critical, the proposed approach represents a promising step toward their future integration into preoperative planning workflows. In spine surgery, CT is often required for accurate assessment of bony anatomy and is commonly registered with intraoperative fluoroscopy or navigation systems. Therefore, the ability to derive CT-like information directly from preoperative MRI could potentially reduce the need for additional CT acquisitions while preserving the anatomical information required for surgical planning and image-guided procedures.

To conclude, our results highlight the potential of rigid-registration-based pseudo-pairing in MR-to-CT translation and establish a strong baseline for further developments toward more spatially coherent and clinically useful synthetic imaging.

\section*{Acknowledgements}
This work has been partially funded by the Basque Government under the HAZITEK 2024 Programme (grant number ZE-2024/00030). Ethical approval and consent requirements were addressed by the original dataset providers. Additionally, ethical approval for the current study was obtained by Comité de Ética de la Investigación con medicamentos de Euskadi (CEIm-E) on October 23, 2025, with code PS2025063.









\printcredits

\bibliographystyle{elsarticle-num-names}

\bibliography{cas-refs-gpt}



\end{document}